\documentclass[12pt,reqno]{article}
\usepackage{graphicx}
\usepackage{amsmath}
\usepackage{times}
\usepackage{hyperref}
\usepackage{float}
\usepackage{accsupp}
\usepackage{fullpage}
\usepackage{subcaption} % For subfigures
\usepackage{caption}
\usepackage{longfigure}
\usepackage{booktabs}
\usepackage{caption}
\begin{document}

\title{
Reasoning-Enhanced Rare-Event Prediction with Balanced Outcome Correction 
}

\author{%
\small
\begin{tabular}{c}
Vitaly Bulgakov$^{1,2}$ and Alexander Turchin$^{1}$\\[1ex]
$^{1}$Mass General Brigham, Boston, MA, USA\\
$^{2}$Profiteya LLC, Boston, MA, USA
\end{tabular}
}

\date{January 22, 2026}

\maketitle

\begin{abstract}
Rare-event prediction is critical in domains such as healthcare, finance, reliability engineering, customer support, aviation safety, where positive outcomes are infrequent yet potentially catastrophic. Extreme class imbalance biases conventional models toward majority-class predictions, limiting recall, calibration, and operational usefulness. 
We propose LPCORP (Low-Prevalence CORrector for Prediction)*, a two-stage framework that combines reasoning-enhanced prediction with confidence-based outcome correction. A reasoning model first produces enriched predictions from narrative inputs, after which a lightweight classifier evaluates and selectively corrects these outputs to mitigate prevalence-driven bias. In this study we used Logistic-Regression (LR) and a simple Multilayer Perceptron (MLP) classifiers for this purpose.

We evaluate LPCORP on real-world datasets from medical and consumer service domains. The results show that this method transforms the original rare-event prediction problem into a more balanced supervised correction task without discarding or resampling observations. Test-set evaluation demonstrates substantially improved performance, particularly in precision, which is a known weakness in low-prevalence data. We further provide a cost-reduction analysis comparing the expenses associated with rare-event damage control without preventive measures to those incurred when low-cost, prediction-based preventive interventions are applied that showed up to 40+\% reduction in some cases.
\\\\
* \textit{Patent pending: U.S. Provisional 63/933,518, filed 8 December 2025.}
\end{abstract}

\section{Introduction}

We introduce LPCORP (Low-Prevalence CORrector for Prediction), a two-stage method that first uses reasoning models to enrich prediction signals and then corrects those reasoning outputs using a conventional machine-learning classifier that reduces bias toward the negative class in rare-event prediction.
\\\\
Rare-event prediction—such as detecting cardiac arrest, ICU deterioration, machine failures, or fraud—presents unique challenges because the positive class represents only a tiny fraction of all cases, yet its consequences are often catastrophic. Although these events are rare, missing them can lead to severe clinical, financial, or safety outcomes. This extreme class imbalance produces noisy signals, sparse supervision, and biased learning dynamics. As a result, conventional models tend to over-predict the majority class.
\\\\
Despite these challenges, rare-event prediction has enormous value. Accurate early detection enables proactive intervention, preventing catastrophic outcomes. In healthcare, even small improvements in prediction performance can translate into substantial clinical and financial benefits: fewer unplanned ICU transfers, reduced in-hospital cardiac arrests, lower readmission penalties, and more efficient resource allocation.
\\\\
A key concept is prevalence, defined as: Prevalence = (\# of positive cases) / (total \# of cases). In rare-event settings, prevalence may be as low as 1–15\%, meaning the model sees far fewer positive examples during training. This imbalance biases the model toward predicting “no event,” and thus requires specialized techniques such as resampling, cost-sensitive learning, class weighting, or threshold tuning to reliably detect the rare cases.
\\\\
In this study, we demonstrate how to overcome the inherent challenges of rare-event prediction and show how improved model design can translate directly into meaningful operational and cost savings. Our datasets consist of narratives, e.g. clinical notes, with labeled outcomes. Below, we highlight real use cases where rare-event prediction drives measurable impact.

\subsubsection*{Use Case 1 — Consumer Finance Complaint (Monetary Relief Prediction)}
In consumer finance, textual descriptions of disputes often contain information that correlates with the likelihood of monetary remediation. Inputs may take the form of free-text complaint narratives or structured textual fields (e.g., Product, Sub-product, Issue, Sub-issue), as provided by the Consumer Financial Protection Bureau (CFPB) complaint database, see \\
https://www.consumerfinance.gov/complaint/, which we used in this study. The prediction task is formulated as a binary classification problem:
\\\\
0 (Negative) — Closed with non-monetary relief
\\
1 (Positive) — Closed with monetary relief
\\\\
Only a small proportion of complaints result in monetary reimbursement, making the positive outcome rare. These cases are operationally and reputationally sensitive, as monetary remediation incurs direct financial costs and can lead to regulatory or customer-satisfaction implications. Accurate early prediction enables more efficient case triage, such as escalation to a specialized resolution team or proactive settlement, reducing time to resolution and financial exposure.

\subsubsection*{Use Case 2 — In-Hospital Cardiac Arrest (IHCA) Prediction}
In clinical settings, patients may experience in-hospital cardiac arrest (IHCA) during admission. IHCA is infrequent—occurring in approximately 1–4\% of adult hospitalizations—yet it is associated with high mortality and substantial resource utilization. Failure to detect early clinical deterioration results in emergency Code Blue activation, intensive care admission with extended length of stay, mechanical ventilation, therapeutic hypothermia, long-term neurological impairment, and potential medicolegal consequences.
For this task, the input consists of sequential clinical notes documenting patient status and clinical reasoning throughout hospitalization. The model predicts a binary outcome:
\\\\
0 (Negative) — No cardiac arrest
\\
1 (Positive) — Cardiac arrest
\\\\
Depending on whether the timestamp of the arrest event is available, the problem can be formulated as either a fixed prediction horizon or a dynamic early warning task. In our research study we consider both cases retrieved from the MIMIC-III clinical dataset and showed here only the one with timestamp available, as it has more practical sense. 

\section{Related work}

In this section, we provide a brief overview of research relevant to rare-event prediction, probability calibration, and reasoning-based prediction. The overview is not intended to be exhaustive; instead, it highlights representative work that illustrates the challenges of low-prevalence prediction and motivates the proposed method across different application domains.
\\\\
\noindent
[1] represents a classic survey explaining why ML models over-predict the majority class and reviewing class weighting, resampling, and cost-sensitive methods.
[2] introduces SMOTE, the foundational method for resampling minority classes in imbalanced datasets.
Empirical proof that ROC AUC (Receiver Operating Characteristic Area Under the Curve) can be misleading when prevalence is low, validating the need for threshold tuning or correction, is given in [3].
[4] shows how model probability outputs are often miscalibrated and need post-processing, especially in imbalanced data.
[5] establishes logistic regression as a strong baseline for clinical prediction.
[6] presents a machine-learning--based system that predicts in-hospital cardiac arrest in the ICU using multimodal data (vitals, labs, and clinical information). The study compares multiple ML models.
[7] reviews applications of artificial intelligence across critical-care medicine---ICU monitoring, deterioration prediction, decision support---and discusses challenges in data sparsity, rare events, model generalization, and clinical deployment.
[8] is dedicated to rare-failure detection with unsupervised learning due to sparse labels.
[9] benchmarks rare-event prediction with real fraud rates ($\approx 0.1\%$).
[10] predicts extremely rare climate anomalies, conceptually similar to rare-event classification.
[11] directly links low-prevalence and extreme events in a market context to biased information flows and demonstrates that standard metrics fail under rare-event conditions.
[12] replaces raw LLM answers with reasoning verification and significantly boosts accuracy on prediction tasks.
[13] represents a survey of how machine learning (ML) and knowledge representation \& reasoning (KRR) intersect, including how reasoning methods can complement ML in low-data, high-uncertainty scenarios.
Finally, a practical tutorial on handling imbalanced classification by Jason Brownlee is provided in [14].

\section{Formalization of the Imbalance Challenge}

A large number of false positives is harmful because every false alarm triggers unnecessary action—such as costly interventions, resource allocation, follow-up investigations, or escalation to experts. In practical terms, false positives waste time and money, overload personnel, reduce trust in the system, and may cause important true positives to be ignored due to alarm fatigue. A standard statistical measure that reflects the proportion of false positives among predicted positives is Precision — high precision means few false positives:

\[
\mathrm{Precision} = \frac{TP}{TP + FP}.
\]

\noindent In rare-event prediction, precision collapses unless the model reduces false positives dramatically—even if accuracy and recall look good. Let:
\begin{itemize}
    \item $G \in \{0,1\}$: the ground-truth label.
    \item $\hat{G} \in \{0,1\}$: the model prediction.
    \item $p = P(G = 1)$: the prevalence of the positive class (rare event), with $p \ll 1$.
\end{itemize}

\[
\mathrm{Accuracy} = P(\hat{G} = G).
\]

\noindent In low-prevalence settings, a trivial classifier (that always predicts one class) predicts the majority class, so

\[
\mathrm{Accuracy}_{\text{trivial}} = P(G = 0) = 1 - p .
\]

\noindent When $p$ is small (e.g., $p = 0.01$ or $1\%$),

\[
\mathrm{Accuracy}_{\text{trivial}} = 0.99 .
\]

\noindent Thus, a model can achieve 99\% accuracy while detecting none of the true positive cases. The metric that reveals this failure is precision:

\[
\mathrm{Precision}
= P(G = 1 \mid \hat{G} = 1)
= \frac{TP}{TP + FP}.
\]

\noindent Because positives are rare, even a small number of false positives may dominate the denominator and precision collapses, even when accuracy is high. More formally, for a fixed true positive rate TPR and false positive rate FPR it is easy to show that

\[
\mathrm{Precision}
= \frac{p \cdot \mathrm{TPR}}
       {p \cdot \mathrm{TPR} + (1 - p)\cdot \mathrm{FPR}} .
\]

\noindent As p → 0

\[
\mathrm{Precision}
\approx \frac{p \cdot \mathrm{TPR}}{(1 - p)\cdot \mathrm{FPR}}
\;\longrightarrow\; 0 .
\]

\noindent unless FPR is extremely small.

\section{Two-stage method}

The method we propose addresses the low-prevalence challenges described above by leveraging a powerful tool — an LLM with reasoning capabilities.

\subsection{Stage 1: Reasoning and conclusion generation with a low-cost LLM}

In the first stage, we collect a dataset of samples, where each sample consists of a textual expression paired with a labeled outcome. The goal of this stage is to generate structured reasoning and a predicted conclusion using a low-cost reasoning language model (LLM). The textual narrative serves as the query to the LLM.
\\
We use \textit{deepseek-ai/DeepSeek-R1-Distill-Llama-8B} [15], a computationally efficient model that inherits the reasoning capabilities of DeepSeek-R1 while leveraging the LLaMA architecture. It can run on small to medium GPU memory and is suitable for execution in a local environment, which is critical for organizations such as medical institutions that handle sensitive data and prefer to avoid external cloud environments for security and compliance reasons. The model is queried using a zero-shot prompt; no training or fine-tuning occurs at this stage.
\\\\
A simplified version of the prompt example is as follows:
\\
--------------------------------------------------------------
\\
\medskip
\noindent\textbf{Task:}  
Predict whether the patient will experience a cardiac arrest (HCA) during hospitalization based on the clinical notes.

\medskip
\noindent\textbf{Notes:}  
\texttt{\{text\}}

\medskip
\noindent\textbf{Instructions:}
\begin{enumerate}
    \item Identify key clinical details in the notes.
    \item Use structured reasoning: for each relevant factor, explain how it supports outcome ``\texttt{\{class0\}}'' or ``\texttt{\{class1\}}''.
\end{enumerate}

\medskip
\noindent\textbf{Final Answer:}  
Return exactly one line:

\medskip
\noindent\texttt{\{class0\}, \{class1\}, or \{not sure\}.}
\\
--------------------------------------------------------------
\\\\
After each LLM execution we have four components:
\\
\begin{table}[H]
\centering
\label{tab:stage1_components}
\begin{tabular}{clll}
\toprule
\textbf{\#} & \textbf{Component} & \textbf{Type} & \textbf{Description} \\
\midrule
1 & Original Text & Input & The raw textual sample (e.g., medical notes). \\
2 & Ground Truth & Input & The known outcome label. \\
3 & Reasoning & Output & The LLM’s step-by-step reasoning text. \\
4 & Conclusion & Output & The LLM’s predicted outcome (class0/class1/not sure). \\
\bottomrule
\end{tabular}
\end{table}

\noindent \textbf{Example for IHCA prediction:}
\\
--------------------------------------------------------------
\\
\textit{Original Text (excerpt):}
\\\\
"Sinus tachycardia. Normal ECG except for rate...
\\
MEDICAL CONDITION: 55-year-old woman with...
\\
The endotracheal tube has been removed. A left subclavian venous line..."
\\\\
\textit{LLM Reasoning (generated excerpt):}
\\\\
The task is to determine whether the patient will experience cardiac arrest (HCA) during hospital stay...
\\
...
\\
The patient shows improvement with treatment. No indication of decompensation leading to HCA...
\\
...
\\
Therefore, the highest probability is that the patient will not have HCA during hospital stay.
\\\\
\textit{LLM Conclusion:} Will not have HCA
\\
\textit{Possible Ground Truth:} Will have HCA
\\
--------------------------------------------------------------
\\\\
In Stage 1 to get reasoning and conclusion we apply just a zero-shot prompt approach to the LLM. The model is not fine-tuned or modified; it simply produces reasoning and conclusions from raw text.

\subsection{Stage 2: Correction with conventional Machine Learning (ML) model}

In the second stage, named Correction, the goal is to determine whether the LLM’s conclusion was correct. The target is binary:
\\\\
1 = LLM conclusion matches the Ground Truth
\\\\
0 = LLM conclusion does not match the Ground Truth
\\\\
This classification task is considerably more balanced than the original rare-event prediction problem, because Correct vs. Incorrect outcomes are typically close to a 50/50 distribution. We use LR and MLP as the Correction Model (CM). The feature matrix X is constructed from TF-IDF text representations (standard libraries) using the concatenation of original text (e.g., clinical notes) + LLM reasoning tex.

We split the dataset into training and test subsets (80\% / 20\%), producing X\_train, X\_test, y\_train, y\_test. The CM model is trained on X\_train / y\_train and evaluated on X\_test / y\_test. The labels in the test set are used solely to compute performance metrics.

For the LLM-generated conclusions corresponding to the test set, we apply correction using the following mapping rule:
\\\\
 % Correction map: \texttt{\{TP $\rightarrow$ FP,\; FP $\rightarrow$ TP,\; TN $\rightarrow$ FN,\; FN $\rightarrow$ TN\}}.
 \textit{If the correction model predicts that the LLM conclusion is incorrect, the predicted class label is flipped. Otherwise, the original prediction is retained.}
\\\\
Effectively, if the CM model predicts the LLM conclusion is incorrect, we flip the outcome (True $\leftrightarrow$ False). If the CM model predicts it is correct, we keep it unchanged. The decision is based on the CM prediction probability and a chosen probability threshold P. We used the out-of-the-box Logistic Regression model and TF-IDF text vectorizer from the scikit-learn package, which have demonstrated strong performance in numerous applications. The vectorizer implements standard n-gram feature extraction, and we used an n-gram range of (1, 2), which we found to be optimal for our study. Although the package provides functionality for extracting the most important features—commonly used in many studies to examine which terms contribute most to the outcome—this was not useful in our case. First, n-grams are usually not explicitly interpretable, and second, our outcome was not event-based but was instead designed to validate the correctness of previous results.

The original prediction task attempts to infer a rare clinical or financial outcome directly from raw narratives. In contrast, the correction model operates on reasoning already produced by the LLM. These reasoning traces often summarize the most relevant evidence, hypotheses, and uncertainty, effectively transforming a complex prediction problem into the simpler task of recognizing whether the reasoning is internally consistent with the final conclusion. Unlike conventional feature engineering, these semantic features are generated automatically by the reasoning model and require no manually designed domain-specific rules. Consequently, the correction problem is substantially easier than predicting the original outcome directly.

\section{Formalization and intuition of the method}

\begin{figure}[H]
\centering
\includegraphics[width=0.8\linewidth]{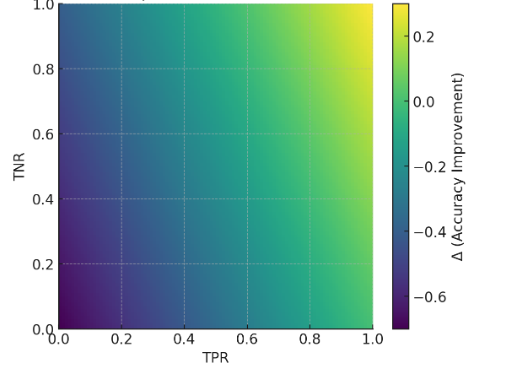}
\caption{Heatmap to show how accuracy net improvement depends on TPR and TNR}
\label{fig:lr_metrics}
\end{figure}

The following mathematical interpretation is a simplified analytical view of how the correction stage improves accuracy. It does not model every nuance of text-based prediction, but it provides an intuitive way to quantify when correction helps and how much improvement can be expected.
\\\\
The following notation has been applied:
\begin{itemize}
    \item Let $G$ be the ground-truth label and $C$ the LLM conclusion (both $\in \{0,1\}$).

    \item Define $S = 1\{C = G\}$ (LLM correctness): $S = 1$ if the LLM was correct, and $S = 0$ otherwise.

    \item Let $\pi = P(S = 1)$ denote the LLM's baseline accuracy (fraction correct).

    \item The CM predicts $\hat{S} \in \{0,1\}$. Define CM operating characteristics:
    \begin{itemize}
        \item $\mathrm{TPR} = P(\hat{S} = 1 \mid S = 1)$ (sensitivity on the ``correct'' class),
        \item $\mathrm{TNR} = P(\hat{S} = 0 \mid S = 0)$ (specificity on the ``incorrect'' class)
    \end{itemize}
\end{itemize}
Here TPR stands for True Positive Rate and TNR for True Negative Rate. Correction rule is:
\\\\
\indent If CM predicts ``incorrect'' ($\hat{S}=0$), flip the LLM conclusion $C \mapsto 1-C$ else keep C
\\\\
Final accuracy after correction is
\[
\begin{aligned}
\mathrm{Acc}_{\mathrm{corr}}
&= P(S = 1)\,P(\hat{S} = 1 \mid S = 1)
   + P(S = 0)\,P(\hat{S} = 0 \mid S = 0) \\
&= \pi \cdot \mathrm{TPR} + (1 - \pi)\cdot \mathrm{TNR}.
\end{aligned}
\]
Let’s do two extreme checks for “Perfect CM” and “Random CM” to confirm this formula:
\begin{itemize}
    \item Perfect CM ($\mathrm{TPR} = \mathrm{TNR} = 1$) $\Rightarrow$ $\mathrm{Acc}_{\mathrm{corr}} = 1$.
    \item Random CM ($\mathrm{TPR} = \mathrm{TNR} = 0.5$) $\Rightarrow$ $\mathrm{Acc}_{\mathrm{corr}} = 0.5$ (regardless of $\pi$).
\end{itemize}
Net improvement is
\[
\Delta = \mathrm{Acc}_{\mathrm{corr}} - \pi
= (1 - \pi)\,\mathrm{TNR} - \pi\,(1 - \mathrm{TPR}).
\]
Correction improves accuracy iff
\[
\Delta > 0, \quad \text{i.e.,} \quad
(1 - \pi)\,\mathrm{TNR} > \pi\,(1 - \mathrm{TPR}).
\]
$\text{If } \pi = 0.7,\ \mathrm{TPR} = 0.8,\ \mathrm{TNR} = 0.75,\ \text{ then}
\;
\mathrm{Acc}_{\mathrm{corr}}
= 0.7 \cdot 0.8 + 0.3 \cdot 0.75
= 0.785$
\\\\
then gain = 0.085
\\\\
Equivalently, for a fixed $\pi$ you need CM to detect incorrect cases well enough relative to its tendency to miss/corrupt correct cases. And one more useful consideration:
\\\\
Under a simplifying assumption that all final errors incur equal cost $c$ (money, harm units), the expected cost per sample before correction is $(1-\pi)c$.
After correction, the expected cost is $(1-\mathrm{Acc}_{\mathrm{corr}})c$.
Therefore, the expected cost reduction per sample is
\[
\text{Savings}
= (\mathrm{Acc}_{\mathrm{corr}} - \pi)\,c
= \Delta \cdot c .
\]
Multiplying by the dataset size $N$, the total expected savings become
\[
N\,\Delta\,c .
\]
To maximize correction efficiency, select the probability threshold P that maximizes the final accuracy:
\[
\mathrm{Acc}_{\mathrm{corr}}
= \pi \cdot \mathrm{TPR}(P) + (1 - \pi)\cdot \mathrm{TNR}(P).
\]
Selection of threshold P can be done experimentally by sweeping its value until the highest accuracy or the lowest cost is achieved:
\[
P_{\text{optimal}} = \arg\max_{P}\,\mathrm{Acc}_{\mathrm{corr}}(P).
\]
or
\[
P_{\text{optimal}} = \arg\min_{P}\,\mathrm{Cost}(P).
\]
The heatmap in Figure \ref{fig:lr_metrics} shows how accuracy net improvement depends on TPR and TNR with base-line LLM accuracy 0.7

\section{Examples}

In these examples, we illustrate the effectiveness of the proposed method by quantifying the cost savings achieved when predictive modeling is used, compared with a baseline in which no prediction is made and no preventive action is taken. This perspective is grounded in prior work. For example, [16] demonstrates how rule-learning techniques can automatically discover behavioral indicators of fraud from large customer-transaction databases.
\\\\
Similarly, [17] establishes how optimal classification decisions should be made when different error types incur different costs. The authors show that, rather than modifying the learning algorithm, one should rely on standard probabilistic classifiers and apply cost-based decision rules that minimize expected loss using a coherent cost matrix. In this view, models provide probability estimates, and decisions are made by mapping these probabilities to actions according to domain-specific costs rather than accuracy-based metrics.
\\\\
Finally, [18] underscores that accurate diagnostic evaluation must account for the unequal consequences of false positives and false negatives. As the authors note, the AUC assigns equal weight to these errors “even though in many applications they have very different consequences,” and that “to make sound decisions, the relative costs of misclassification must be taken into account.” Following this principle, we adopt a cost-minimization framework that balances the high cost of missed clinical events (ce) against the lower, but nonzero, cost of preventive interventions (ci), adjusted for intervention efficacy and event prevalence.
\\\\
Our cost formula can be easily derived based on the following notation:
\begin{itemize}

\item $ep$: event prevalence; probability that the adverse event occurs.

\item $R$: recall (sensitivity); probability the model correctly identifies an event when it occurs,
\[
R = \frac{TP}{TP + FN}.
\]

\item $Pr$: precision (positive predictive value); probability that a predicted event is truly positive,
\[
Pr = \frac{TP}{TP + FP}.
\]

\item $c_e$: event cost; cost incurred if the event occurs and is not prevented.

\item $c_i$: intervention cost; cost of applying a preventive intervention to predicted positives.

\item $e$: intervention efficacy; probability that the intervention prevents the event when applied to a true positive.

\medskip

\item $TP = ep \cdot R$: probability mass of true positives.

\item $FP = TP\left(\frac{1}{Pr} - 1\right)$: probability mass of false positives.

\item $FN = ep(1 - R)$: probability mass of false negatives.

\item $TP + FP = \frac{TP}{Pr}$: number of interventions applied (in expectation).

\end{itemize}
Then it will be
\[
\mathrm{Cost}
= ep \cdot \left(
R\left(\frac{c_i}{Pr} - e \cdot c_e\right) + c_e
\right).
\]
This formula is a compact, closed-form version of standard expected-loss expressions under the assumptions that interventions only affect true positives. It is rewritten in terms of Precision, Recall, and Prevalence instead of raw confusion-matrix counts. We also use a cost reduction (or Return) formula to show effectiveness of the prediction method:
\\
Let
\[
C_{\text{baseline}} = ep \cdot c_e
\]
be the expected cost without any prediction model, and let $C_{\text{model}}$ be the expected cost when using the model.
\\
Then

\[
\text{Return \%}
= \frac{C_{\text{baseline}} - C_{\text{model}}}{C_{\text{baseline}}}
\times 100 .
\]
Whereas event prevalence is calculated directly from the dataset, event cost and intervention cost must be specified externally based on domain knowledge, published estimates, or scenario-based assumptions. Obtaining accurate values for these parameters typically requires dedicated economic or clinical analysis for the specific use case. 

One way to estimate the cost reduction without obtaining accurate treatment and other costs is to use the cost ratio $k = c_i / c_e$, where $c_i$ is the intervention cost and $c_e$ is the event cost introduced earlier. It is much easier to estimate this ratio than the absolute cost and in most cases it should be sufficient for evaluation of method's performance. Starting from the return definition given earlier it is easy to obtain the following formula for the return 
\[
\mathrm{Return} = R \left(e - \frac{k}{Pr}\right).
\]

\subsection{Experimental parameters}
The following settings have been used in the numerical experiments:
\begin{table}[ht]
\centering
\label{tab:experimental_settings}
\begin{tabular}{ll}
\hline
\textbf{Parameter} & \textbf{Value} \\
\hline
Reasoning Model & \texttt{deepseek-ai/DeepSeek-R1-Distill-Llama-8B} \\
Correction Models & LR, MLP \\
Correction Model n-gram Range & (1, 2) \\
Train/Test Split & 0.8 / 0.2 \\
Intervention-to-Event Cost Ratio & 0.02 \\
Intervention Efficacy & 1.0 \\
Correction Probability Threshold & 0.5, 0.6, 0.7, 0.8, 0.9 \\
Max text features in correction & 5000 \\
\hline
\end{tabular}
\end{table}

\noindent For each dataset, we present the following four sets of results:
\begin{enumerate}
    \item Test set evaluation metrics for logistic regression (LR) correction across different correction probability thresholds.
    \item Test set evaluation metrics for multilayer perceptron (MLP) correction across different correction probability thresholds.
    \item Comparison of return (cost reduction) achieved by LR correction, MLP correction, and the baseline Reasoning Only approach.
    \item Summary table highlighting the configuration that achieves the highest cost reduction.
\end{enumerate}

\subsection{Example 1. Consumer Finance Complaint (Monetary Relief Prediction)}
This example is described in Use Case 1 section. We used the dataset provided by Financial Protection Bureau (CFPB) complaint database.
Handling consumer finance complaints that lead to monetary relief is costly both for firms and the overall financial ecosystem: empirical studies of CFPB complaint data show an average payout of about \$1,470 per successful complaint, reflecting the direct financial burden of remediation for companies and consumers alike [19]. At the system level, the Consumer Financial Protection Bureau’s enforcement and supervision work has generated nearly \$20 billion in consumer relief, illustrating the magnitude of unresolved harm that must eventually be addressed without early intervention [20]. By contrast, prevention and support efforts—such as regulatory complaint systems and proactive compliance—are funded through agency operations that cost hundreds of millions annually (e.g., CFPB spending well below \$1 billion per year) and aim to reduce complaint volume and remediation costs over time. Together these figures underscore that proactive prevention and early prediction models are significantly more cost-efficient than reactive handling of complaint events, making them valuable investments for minimizing financial and operational exposure in consumer finance.
\\
A reasonable estimate for the intervention cost in consumer finance complaint handling is \$50–\$150 per flagged complaint, corresponding to 1–3 additional customer interactions (e.g., outbound call, follow-up, documentation) at approximately \$10–\$50 per contact, or equivalently 1–3 hours of agent or specialist time at \$26–\$30 per hour. These ranges are consistent with published benchmarks for financial-services contact center costs and staffing rates, e.g. [21].
\\\\
\noindent With this dataset we had the following sample counts:
\begin{table}[ht]
\centering
\begin{tabular}{lc}
\hline
\textbf{Statistic} & \textbf{Value} \\
\hline
Total samples & 9,530 \\
Positive samples & 2,046 \\
Negative samples & 7,484 \\
Prevalence & 21.47\% \\
\hline
\end{tabular}
\end{table}

\begin{figure}[H]
\centering
\includegraphics[scale=0.6]{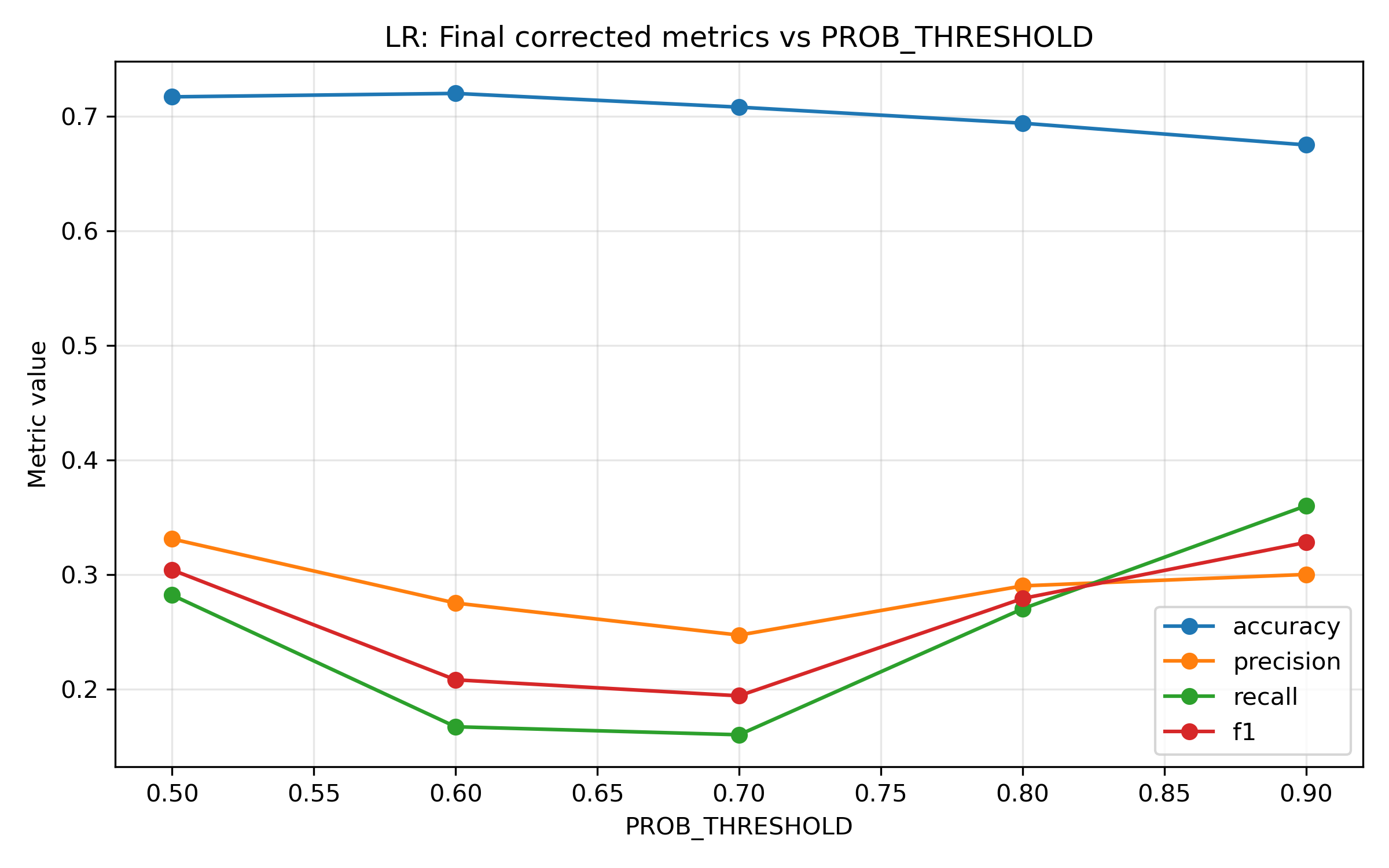}
\caption{Consumer Finance Complaint. Evaluation metrics for LR correction across different correction
probability thresholds}
 \label{fig:LR_metrics_vs_threshold_consumer_finance_complaint}
\end{figure}

\begin{figure}[H]
\centering
\includegraphics[scale=0.6]{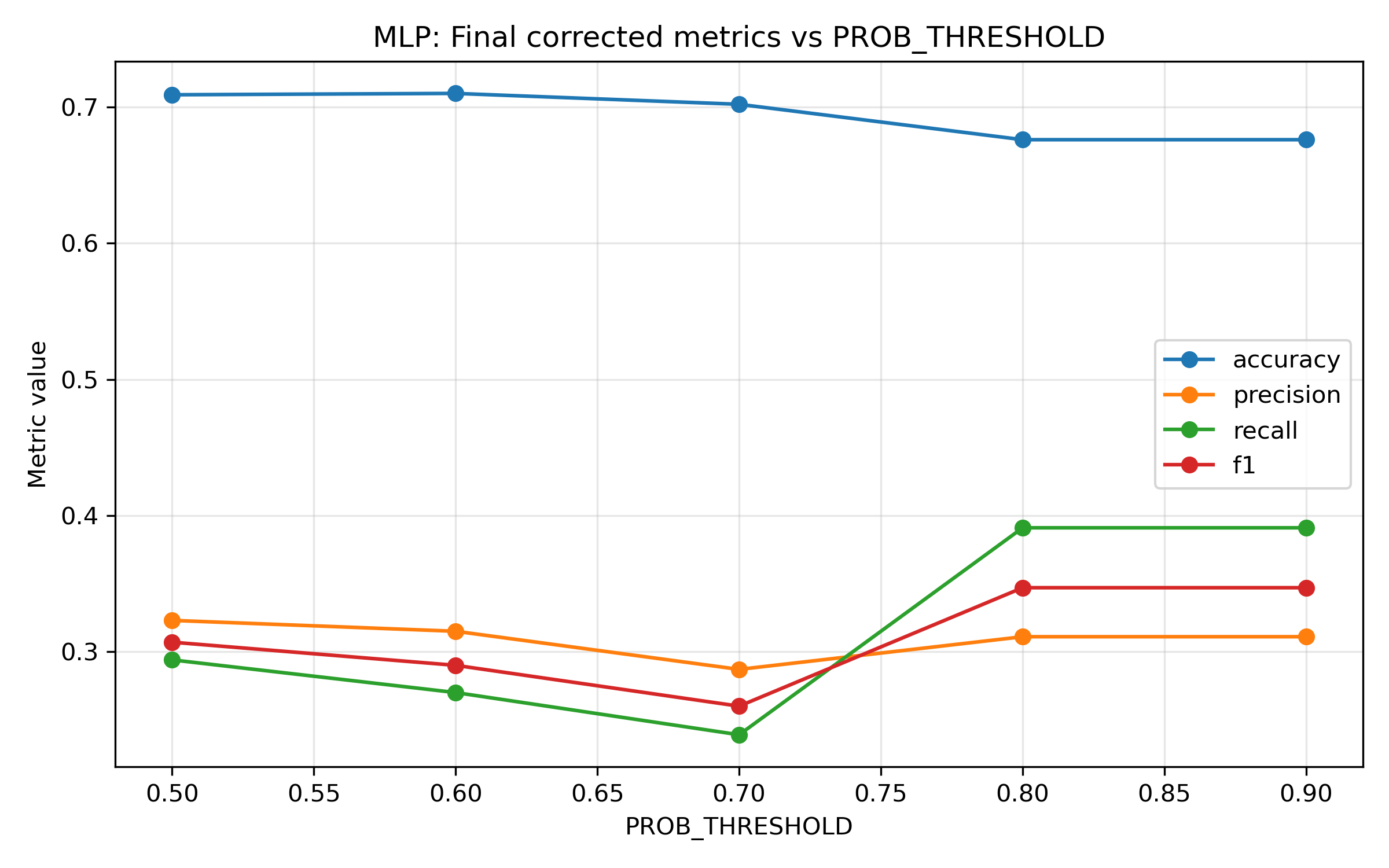}
\caption{Consumer Finance Complaint. Evaluation metrics for MLP correction across different correction
probability thresholds}
 \label{fig:MLP_metrics_vs_threshold_consumer_finance_complaint}
\end{figure}

\begin{figure}[H]
\centering
\includegraphics[scale=0.6]{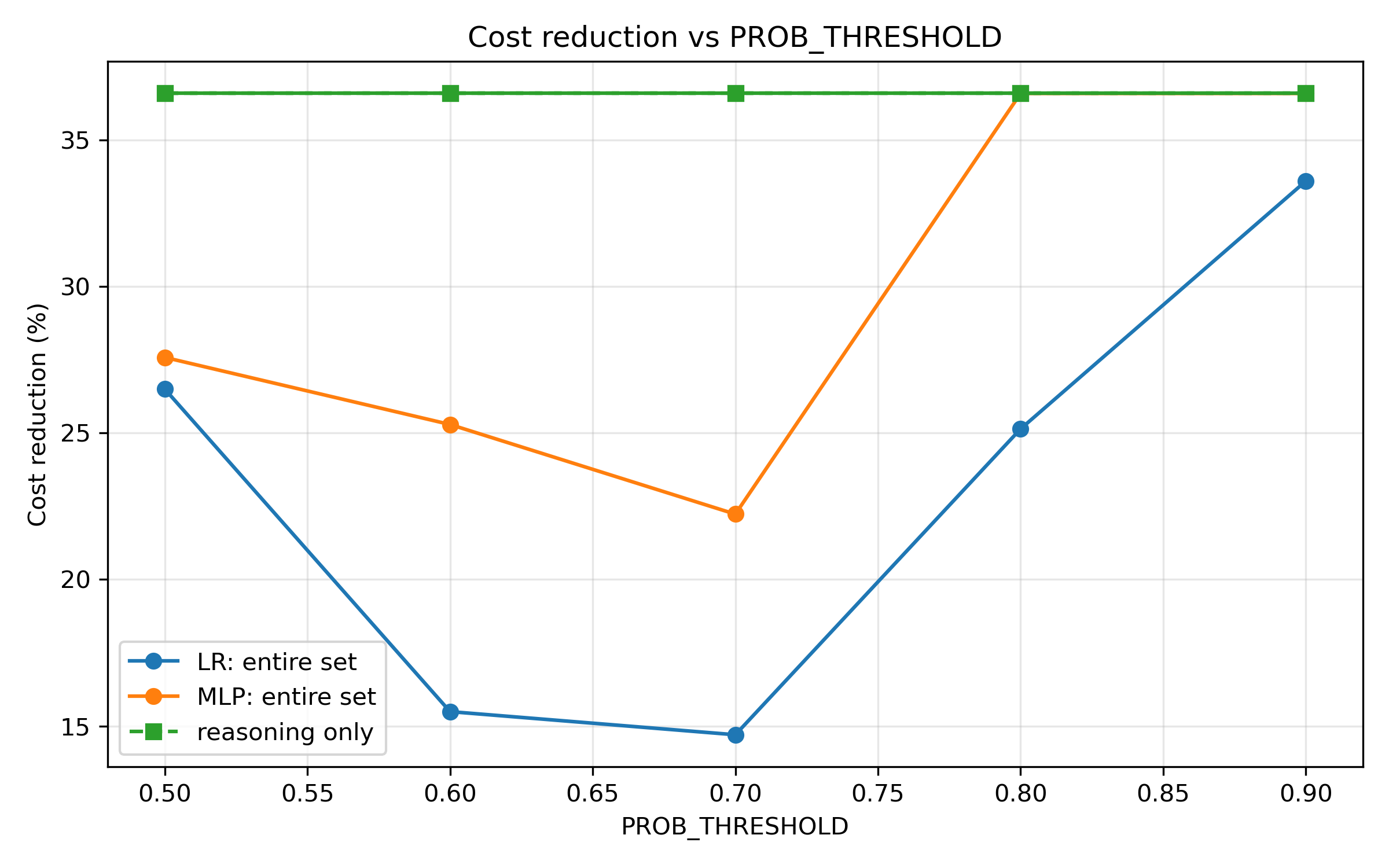}
\caption{Consumer Finance Complaint. Comparison of return (cost reduction) achieved by LR correction, MLP correction, and the
baseline Reasoning Only approach}
 \label{fig:LR_MLP_costs_vs_threshold_consumer_finance_complaint}
\end{figure}

\begin{figure}[H]
\centering
\includegraphics[scale=0.9]{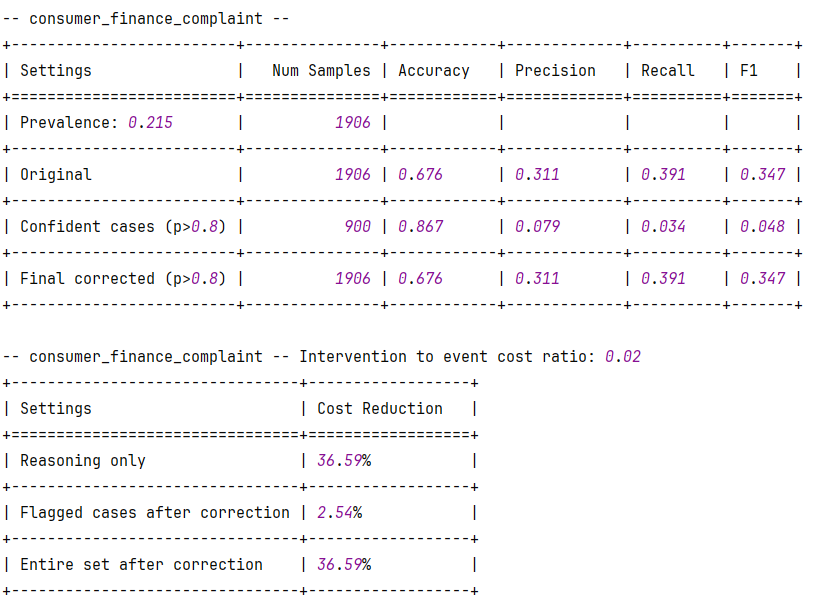}
\caption{Consumer Finance Complaint. Configuration that achieves the highest cost reduction}
 \label{fig:consumer_finance_complaint_best_Prob_Thresh_MLP}
\end{figure}

Figures \ref{fig:LR_metrics_vs_threshold_consumer_finance_complaint} and \ref{fig:MLP_metrics_vs_threshold_consumer_finance_complaint} demonstrate how prediction metrics change with different probability threshold values for both Correction Models (CM). Figure \ref{fig:LR_MLP_costs_vs_threshold_consumer_finance_complaint} demonstrates that among the correction models, MLP achieved the highest return at a probability threshold of 0.8. However, neither correction model outperformed the Reasoning Only baseline. Tables in Figure  \ref{fig:consumer_finance_complaint_best_Prob_Thresh_MLP} confirm this conclusion.
\\

\subsection{Example 2. Prediction of hospital readmission within 30 days}
This example addresses the task of predicting hospital readmission within 30 days after discharge. The data were collected from the MIMIC-IV dataset available through physionet.org. Admission labels were obtained from the ADMISSIONS.csv table in MIMIC-IV Core, and discharge summaries were taken from the DISCHARGE.csv table in MIMIC-IV Notes. After preprocessing, we randomly selected 10,000 admission-level samples (each corresponding to a single hospital stay). After running the reasoning process on all samples, a small number of cases were excluded because no definitive conclusion could be reached. The cost-saving evaluation was then performed on the remaining test samples.

Published U.S. data consistently show that hospital readmissions are costly, while prevention and care-transition interventions are comparatively inexpensive, even when implemented broadly. National Healthcare Cost and Utilization Project (HCUP) [22] analyses estimate that the average cost of an all-cause 30-day adult hospital readmission was approximately \$15,200 in 2018 and increased to about \$16,300 by 2020, representing costs that are roughly 12\% higher than the original (index) admission. Independent literature synthesis supports these figures: a meta-analysis pooling estimates across multiple studies reports a mean readmission cost of approximately \$16,870 per event. Taken together, these sources support using \$15,000–\$17,000 per readmission as a defensible modeling range for U.S. hospitals and payers.
In contrast, the cost of readmission-prevention and care-transition programs is substantially lower. Published randomized and observational studies suggest that structured transitional care services cost on the order of \$500–\$600 per patient per year. Medicare’s Transitional Care Management (TCM) benefit reimburses providers approximately \$236 per care-transition episode, reflecting the payer-level cost of a standardized prevention intervention. Other nurse-led or “transition coach”–style programs report per-case costs of roughly \$500, often alongside measurable reductions in readmissions.

Overall, while exact costs vary by population, condition, and implementation model, the literature consistently shows that preventive interventions costing a few hundred dollars per patient can offset or avert readmissions costing tens of thousands of dollars, highlighting the strong economic rationale for early identification and prevention strategies. 

\noindent With this dataset we had the following sample counts:
\begin{table}[ht]
\centering
\begin{tabular}{lc}
\hline
\textbf{Statistic} & \textbf{Value} \\
\hline
Total samples & 9,209 \\
Positive samples & 2,041 \\
Negative samples & 7,168 \\
Prevalence & 22.16\% \\
\hline
\end{tabular}
\end{table}

\begin{figure}[H]
\centering
\includegraphics[scale=0.6]{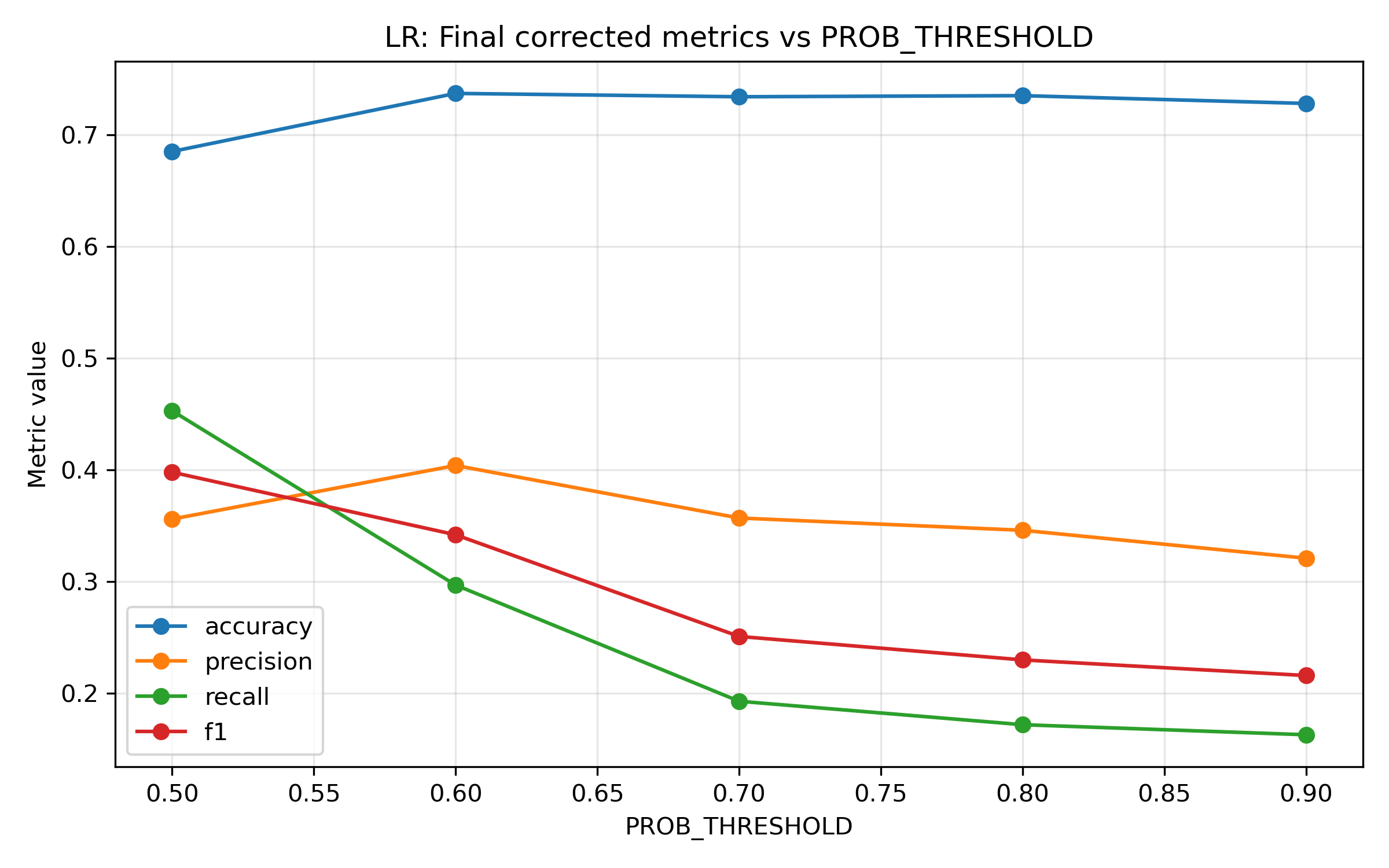}
\caption{Readmission within 30 days. Evaluation metrics for LR correction across different correction
probability thresholds}
 \label{fig:LR_metrics_vs_threshold_mimic_4_discharge}
\end{figure}

\begin{figure}[H]
\centering
\includegraphics[scale=0.6]{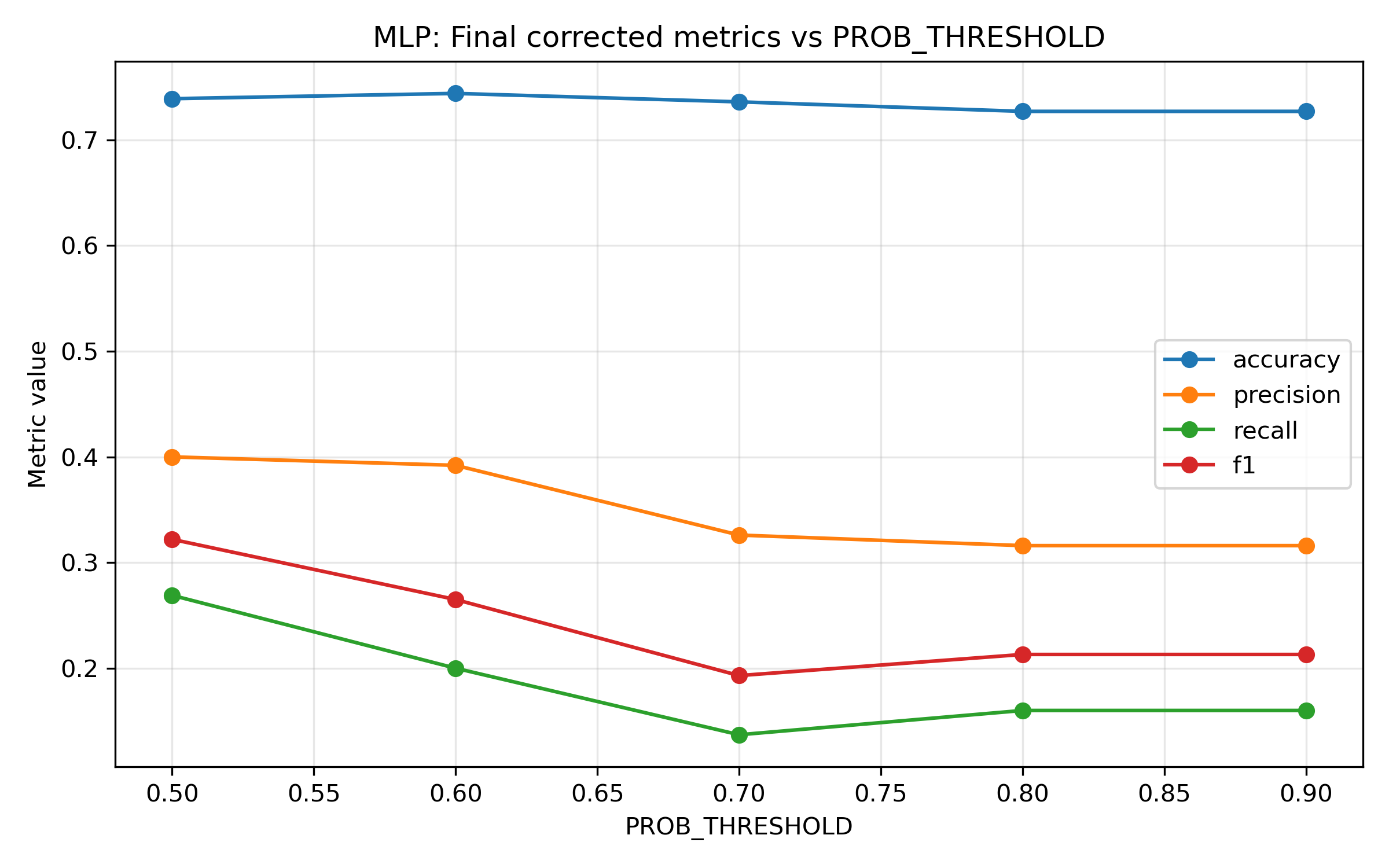}
\caption{Readmission within 30 days. Evaluation metrics for MLP correction across different correction
probability thresholds}
 \label{fig:MLP_metrics_vs_threshold_mimic_4_discharge}
\end{figure}

\begin{figure}[H]
\centering
\includegraphics[scale=0.6]{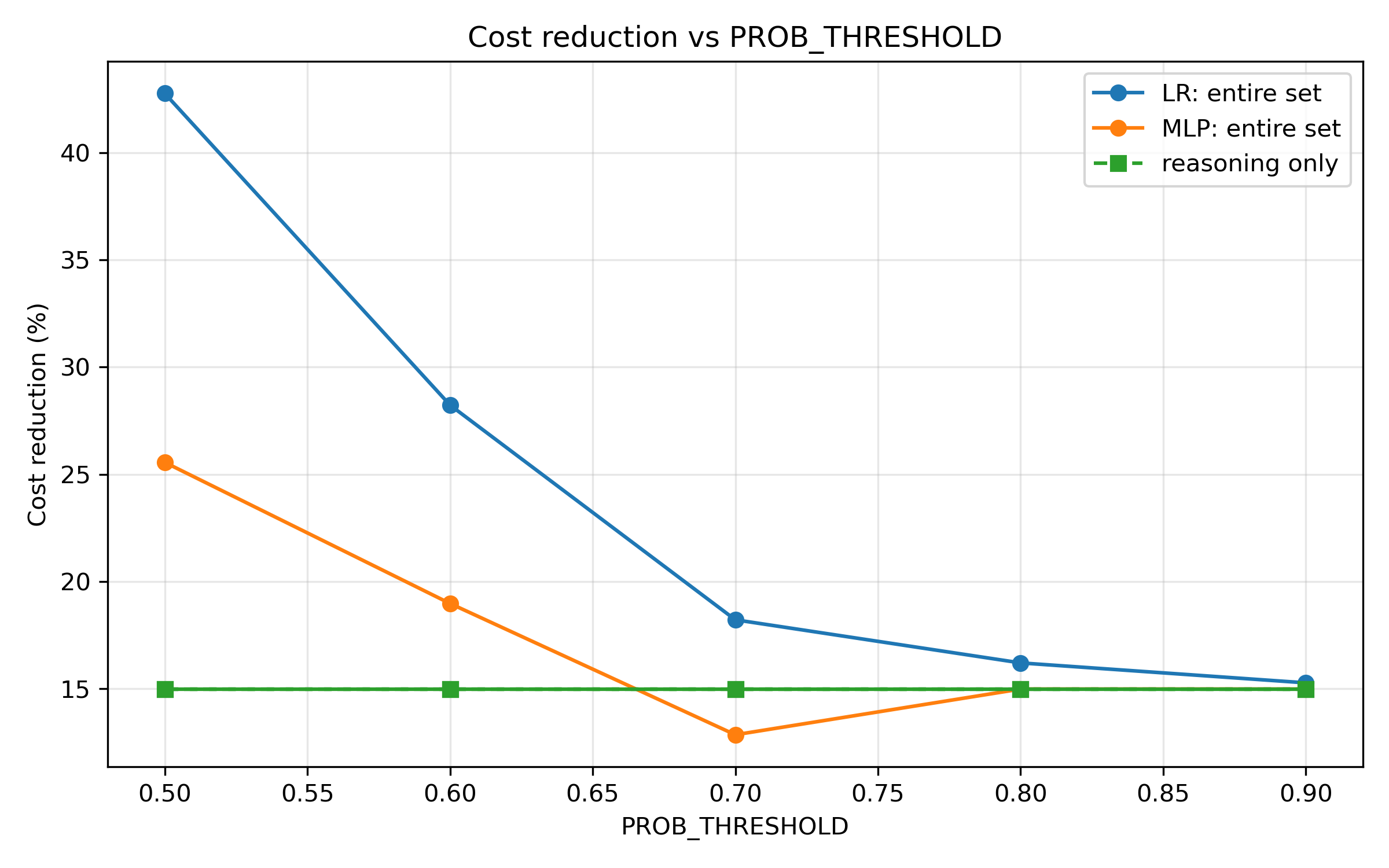}
\caption{Readmission within 30 days. Comparison of return (cost reduction) achieved by LR correction, MLP correction, and the
baseline Reasoning Only approach}
 \label{fig:LR_MLP_costs_vs_threshold_mimic_4_discharge}
\end{figure}

\begin{figure}[H]
\centering
\includegraphics[scale=0.9]{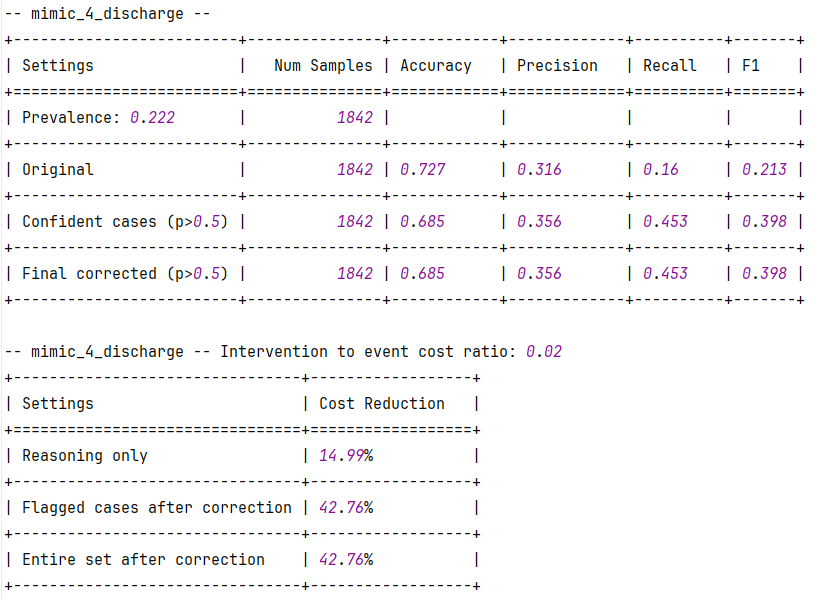}
\caption{Readmission within 30 days. Configuration that achieves the highest cost reduction}
 \label{fig:mimic_4_discharge_best_Prob_Thresh_LR}
\end{figure}

Figures \ref{fig:LR_metrics_vs_threshold_mimic_4_discharge} and \ref{fig:MLP_metrics_vs_threshold_mimic_4_discharge} demonstrate how prediction metrics change with different probability threshold values for both Correction Models (CM). Figure \ref{fig:LR_MLP_costs_vs_threshold_mimic_4_discharge} demonstrates that LR CM substantially outperforms both Reasoning Only and MLP corrected predictions and achieves the best result in terms of cost reduction with probability threshold 0.5. Tables in Figure \ref{fig:mimic_4_discharge_best_Prob_Thresh_LR} demonstrates the effect of LR correction.
\\

\subsection{Example 3. Timestamped IHCA}

\subsubsection{Data Retrieval}

As mentioned earlier, the data for this experiment were obtained from the MIMIC-III dataset available through PhysioNet. The source tables processed to produce the final labeled clinical note samples were D\_ITEMS.csv, PROCEDUREEVENTS\_MV.csv, NOTEEVENTS.csv, ADMISSIONS.csv, and PATIENTS.csv. As in the previous example, after preprocessing, we randomly selected 10,000 admission-level samples (each corresponding to a single hospital stay with associated clinical notes and events). After running the reasoning process on all samples, a small number of cases were excluded because no definitive conclusion could be reached. The cost-saving evaluation was then performed on the remaining test samples.

This example focuses on In-Hospital Cardiac Arrest (IHCA) prediction. IHCA can be identified in the MIMIC database using two fundamentally different approaches, which differ in clinical meaning, temporal precision, and suitability for predictive modeling. They are Timestamped IHCA (event-based identification) and ICD-coded IHCA. But only the first one allows for modeling the real use case to predict the cardiac arrest within the next H hours, where H is a selected window. We took H=6, as recommended in multiple studies.

Timestamped IHCA (event-based identification) reflects actual, time-specific cardiac arrest events documented during clinical care. This enables time-to-event modeling (e.g., “predict arrest within the next 6 hours”), but captures only arrests where CPR/defibrillation was documented resulting in a relatively small number of positive cases. To prevent label leakage and ensure a realistic prospective prediction setup, we followed the outcome-blind window (prediction gap) design widely used in clinical deterioration and early-warning literature. In this framework, all patient data occurring within a fixed time horizon immediately preceding the outcome are excluded from model inputs (McDermott et al., 2021 [23]). Prior work has applied such exclusion windows ranging from 2 to 12 hours to avoid models learning from charting patterns or physiological measurements that occur after clinicians have already recognized impending deterioration (e.g., ICU transfer and AKI prediction studies [24]).

% Our data retrieval and cleaning resulted in prevalence (\% of positive cases) of 1-2\%. This number matches various medical sources. For example, In‑Hospital Cardiac Arrest: A Review (Andersen LW et al., 2019) [21] – This review reports that national cardiac-arrest registries list rates of IHCA between 1.2 and 10 per 1000 admissions (0.12\%–1.0\%).

Our data retrieval and cleaning resulted in an IHCA prevalence of approximately 2\%. This value is within the range reported in clinical studies [25], [26], depending on population and case-definition. For example, Andersen et al. (2019) report that national cardiac arrest registries observe 1.2 to 10 events per 1,000 hospital admissions ($\approx 0.12\%\text{--}1.0\%$) for timestamped, resuscitation-documented IHCA events [21]. Higher prevalence values are often observed when broader case-definitions are used (e.g., ICD-coded arrests or ICU-only cohorts), which include arrests without documented resuscitation time.
% As the number of positive samples in this case was limited, we applied down-sampling to the majority (negative) class.

\subsubsection{Notes selection for positive and negative cases}

For IHCA-positive admissions, clinical notes are collected from the time of hospital admission up to a fixed prediction cutoff defined as 6 hours before the first documented in-hospital cardiac arrest (IHCA). This ensures that only information available prior to the prediction horizon is used for modeling.

For IHCA-negative admissions, where no true IHCA time exists, a pseudo-event time is assigned. This pseudo-event time is sampled from the empirical distribution of time-since-admission observed in IHCA-positive cases, subject to the constraint that it lies within the admission’s length of stay. Notes for negative admissions are collected using the same temporal rule as for positives, up to 6 hours before the pseudo-event time. This approach aligns the observation windows of positive and negative cases and prevents temporal bias.

Negative pseudo-event times are randomly sampled to match the time-since-admission distribution of IHCA-positive events. This can be expressed with the following formulas:

\subsubsection*{Index Time Construction and Note Window Definition}

Let
\begin{itemize}
    \item $T_{\text{adm}}$: admission time.
    \item $T_{\text{dis}}$: discharge time.
    \item $T_{\text{IHCA}}$: time of first IHCA (positive admissions only).
    \item $H=6$ hours: prediction horizon.
    \item $\mathrm{LOS}=T_{\text{dis}}-T_{\text{adm}}$: length of stay.
\end{itemize}

\paragraph{Positive Admissions.}
For IHCA-positive admissions, the index (prediction) time is defined as

\[
t^{+}_{\text{index}} = T_{\text{IHCA}} - H.
\]

Only notes available prior to the prediction time are included:

\[
T_{\text{adm}} \le T_{\text{note}} \le t^{+}_{\text{index}}.
\]

Let

\[
\Delta t^{+} = T_{\text{IHCA}} - T_{\text{adm}}
\]

denote the observed time from admission to IHCA.

\paragraph{Negative Admissions.}
For IHCA-negative admissions, a pseudo-event time is sampled in order to match the temporal structure of positive cases. Specifically, a pseudo-event offset is drawn from the empirical distribution of positive offsets, subject to a feasibility constraint:

\[
\Delta t^{-} \sim \Delta t^{+}\ \big|\ \Delta t^{+} \le \mathrm{LOS}-H .
\]

Equivalently, this ensures that the corresponding index time satisfies

\[
t^{-}_{\text{index}} = T_{\text{adm}} + \Delta t^{-} \le T_{\text{dis}} - H.
\]

Only notes available prior to this index time are included:

\[
T_{\text{adm}} \le T_{\text{note}} \le t^{-}_{\text{index}}.
\]

\paragraph{Definitions.}
\begin{itemize}
    \item $t^{+}_{\text{index}}$, $t^{-}_{\text{index}}$: cutoff times for note collection in positive and negative admissions.
    \item $\Delta t^{+}$: observed time from admission to IHCA in positive cases.
    \item $\Delta t^{-}$: sampled pseudo-event time from the truncated positive distribution.
    \item $T_{\text{note}}$: timestamp of an individual clinical note.
\end{itemize}

\begin{figure}[H]
\centering
\includegraphics[scale=0.5]{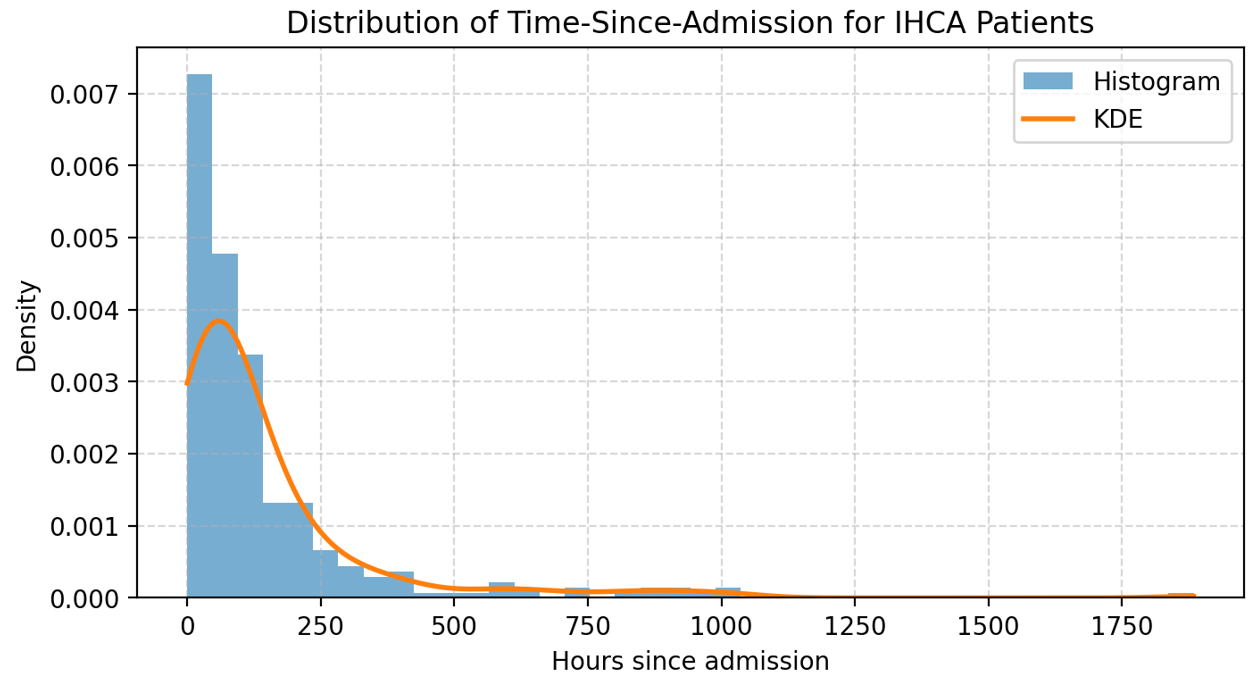}
\caption{Empirical distribution of time since hospital admission to first in-hospital cardiac arrest (IHCA) with a kernel density estimate (KDE) of the same distribution.}
 \label{fig:figures/fig4}
\end{figure}

\begin{figure}[H]
\centering
\includegraphics[scale=0.5]{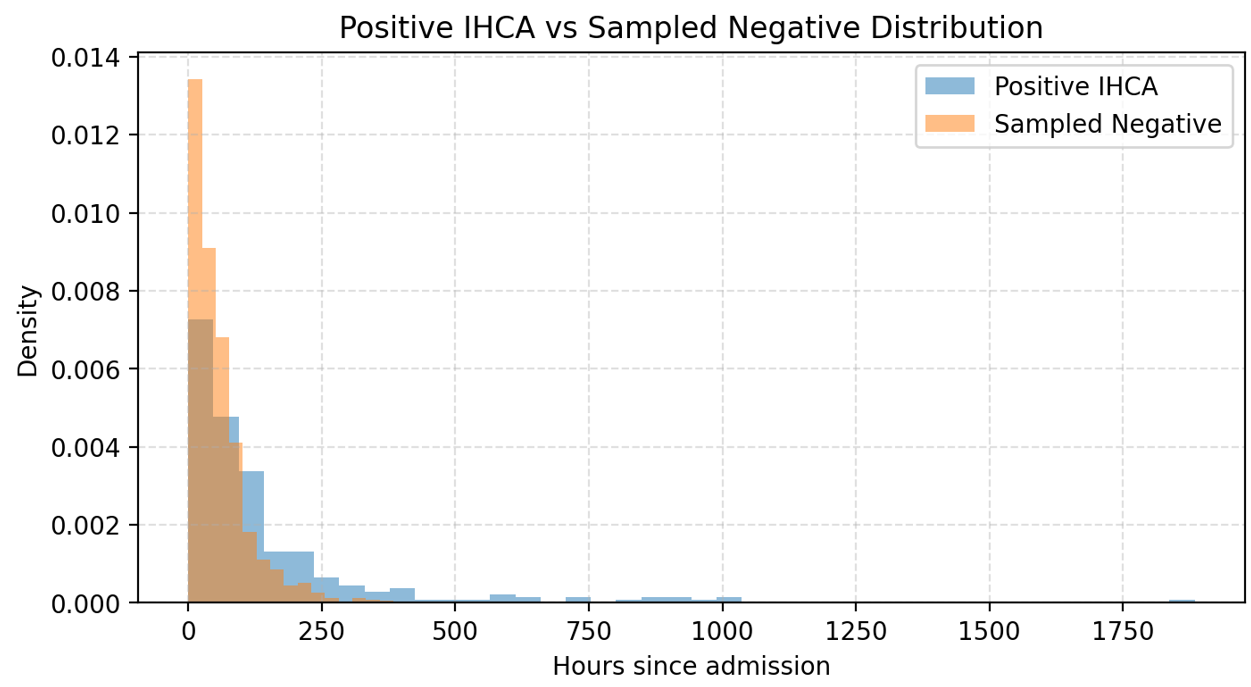}
\caption{Compares the distribution of time since hospital admission for IHCA-positive admissions (blue) and IHCA-negative admissions with sampled pseudo-event times (orange)}
 \label{figures/fig23}
\end{figure}

\paragraph{Interpretation.}
This construction aligns positive and negative admissions at comparable prediction horizons with identical note-availability constraints relative to the index time, while preventing negative admissions from using future information beyond discharge.
\\\\
Figures \ref{fig:figures/fig4} and \ref{figures/fig23} demonstrate graphs of the distribution of time-since-admission to IHCA for positive cases and the corresponding sampled pseudo-event times for negative cases, demonstrating close alignment between the two distributions and validating the matched sampling strategy.

\noindent Figure \ref{fig:figures/fig4} shows the empirical distribution of time since hospital admission to first in-hospital cardiac arrest (IHCA) among IHCA-positive admissions. The blue bars represent the normalized histogram of observed times (in hours), while the orange curve shows a kernel density estimate (KDE) of the same distribution. The distribution is strongly right-skewed, with the highest density occurring early after admission, indicating that a substantial proportion of IHCAs occur within the first days of hospitalization. At the same time, the long right tail demonstrates that cardiac arrest can also occur much later during prolonged hospital stays, extending to several weeks. This empirical time-since-admission distribution is used as a reference distribution for sampling pseudo-event times in IHCA-negative admissions, ensuring that negative cases are temporally aligned with positives when defining prediction cutoffs and note collection windows.
Figure \ref{figures/fig23} is targeted to confirm that the sampling strategy successfully aligns positive and negative admissions. It compares the distribution of time since hospital admission for IHCA-positive admissions (blue) and IHCA-negative admissions with sampled pseudo-event times (orange). The blue histogram represents the empirical distribution of time-to-first IHCA among positive cases. The orange histogram shows the distribution of pseudo-event times assigned to negative cases, which are sampled from the positive IHCA time-since-admission distribution and constrained by each admission’s length of stay. The substantial overlap between the two distributions demonstrates that the negative pseudo-event times closely mirror the timing of true IHCA events. This shows that the sampling strategy successfully aligns positive and negative admissions in terms of time since admission, ensuring comparable observation windows and reducing temporal bias when defining prediction cutoffs and collecting clinical notes.

\subsubsection{IHCA treatment related costs}

Based on analysis of published sources reporting index hospitalization costs, post-arrest ICU care, and specialized resuscitation treatments, we estimate the average cost of IHCA treatment to be approximately \$30,000–\$70,000 per case. We retrieved an approximate IHCA cost from the following sources. Chan et al. (2014) [27] – Reported that the mean cost of the index hospitalization after in-hospital cardiac arrest was about \$35,808 per patient capturing total inpatient costs for IHCA survivors. Damluji et al. (2018) [28] – Designed to estimate costs of index hospitalizations after cardiac arrest in the U.S. using Nationwide Inpatient Sample data—a key reference for hospitalization cost analyses. Nanjayya et al. (2024) [29] – Found mean ICU and hospital costs in cardiac arrest patients with survivors having ICU costs around \$53,735 and overall hospital costs around \$33,048—supporting the tens-of-thousands range.

Published evidence on costs per signal-triggered rapid response activation is limited. Annual MET operating costs have been reported (Bonafide et al. [30]), and typical RRT (Rapid Response Team) activation frequencies are described in systematic reviews (McGaughey et al. [31]), which allow staffing cost per activation to be projected in the hundreds of dollars. Broader discussions of rapid-response team economics also appear in the general literature (e.g., Jones et al., 2011 [32]), though specific micro-costing per activation is not yet well established in peer-reviewed journals.

\subsubsection{Results}

\noindent With this dataset we had the following sample counts:
\begin{table}[ht]
\centering
\begin{tabular}{lc}
\hline
\textbf{Statistic} & \textbf{Value} \\
\hline
Total samples & 8,931 \\
Positive samples & 205 \\
Negative samples & 8,726 \\
Prevalence & 2.3\% \\
\hline
\end{tabular}
\end{table}

\begin{figure}[H]
\centering
\includegraphics[scale=0.6]{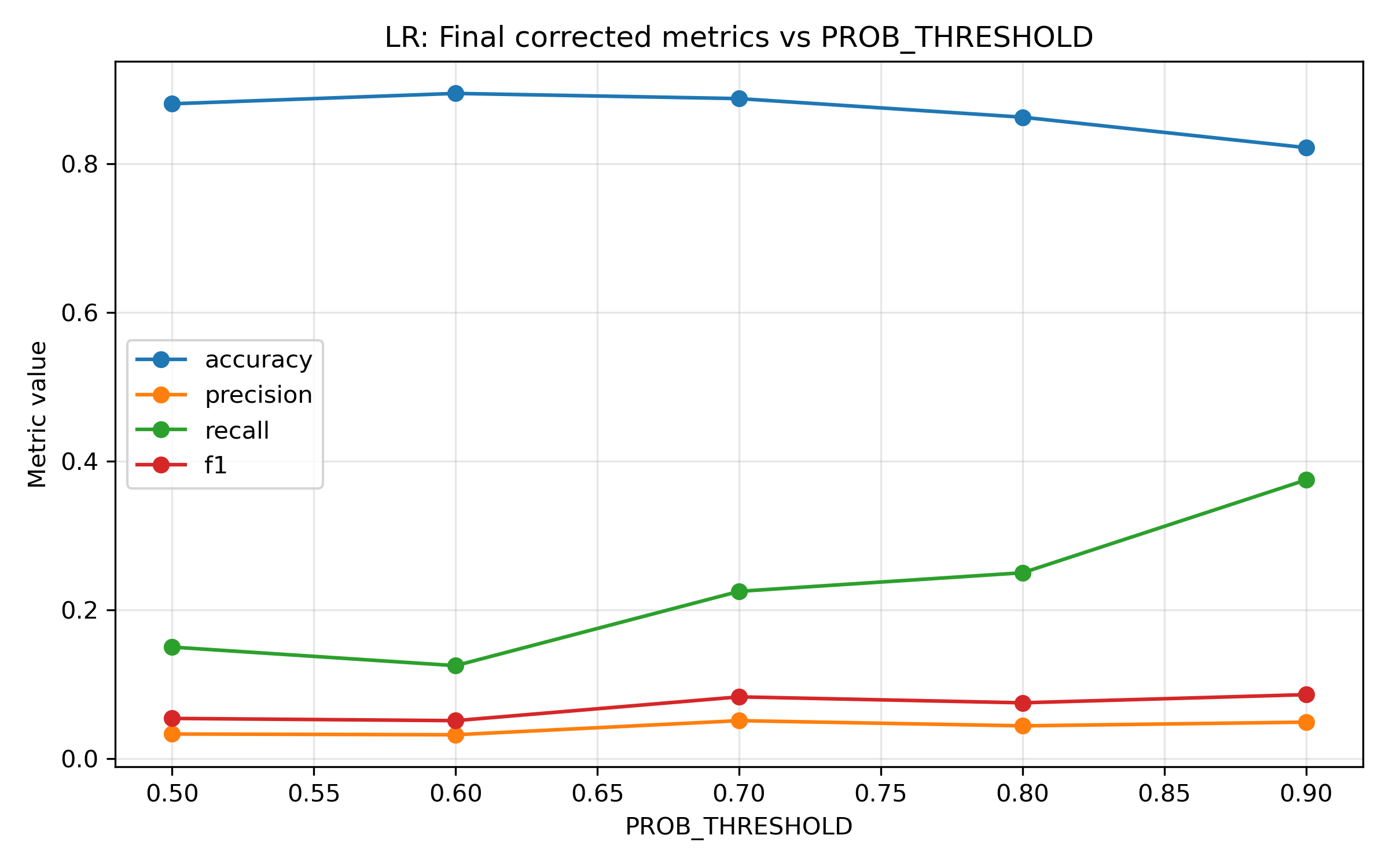}
\caption{IHCA prediction. Evaluation metrics for LR correction across different correction
probability thresholds}
 \label{fig:LR_metrics_vs_threshold_mimic_3_timestamped_HCA}
\end{figure}

\begin{figure}[H]
\centering
\includegraphics[scale=0.6]{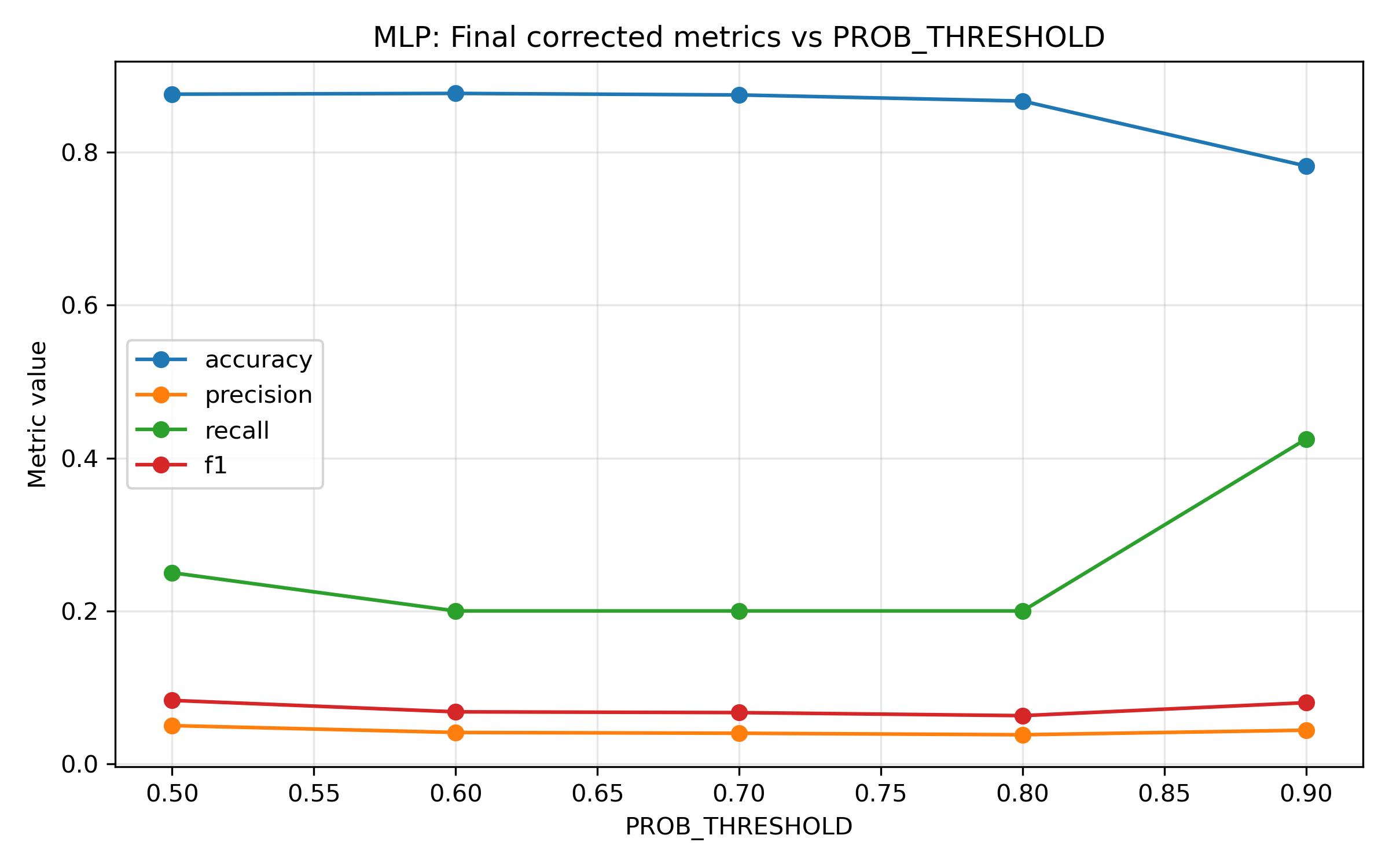}
\caption{IHCA prediction. Evaluation metrics for MLP correction across different correction
probability thresholds}
 \label{fig:MLP_metrics_vs_threshold_mimic_3_timestamped_HCA}
\end{figure}

\begin{figure}[H]
\centering
\includegraphics[scale=0.6]{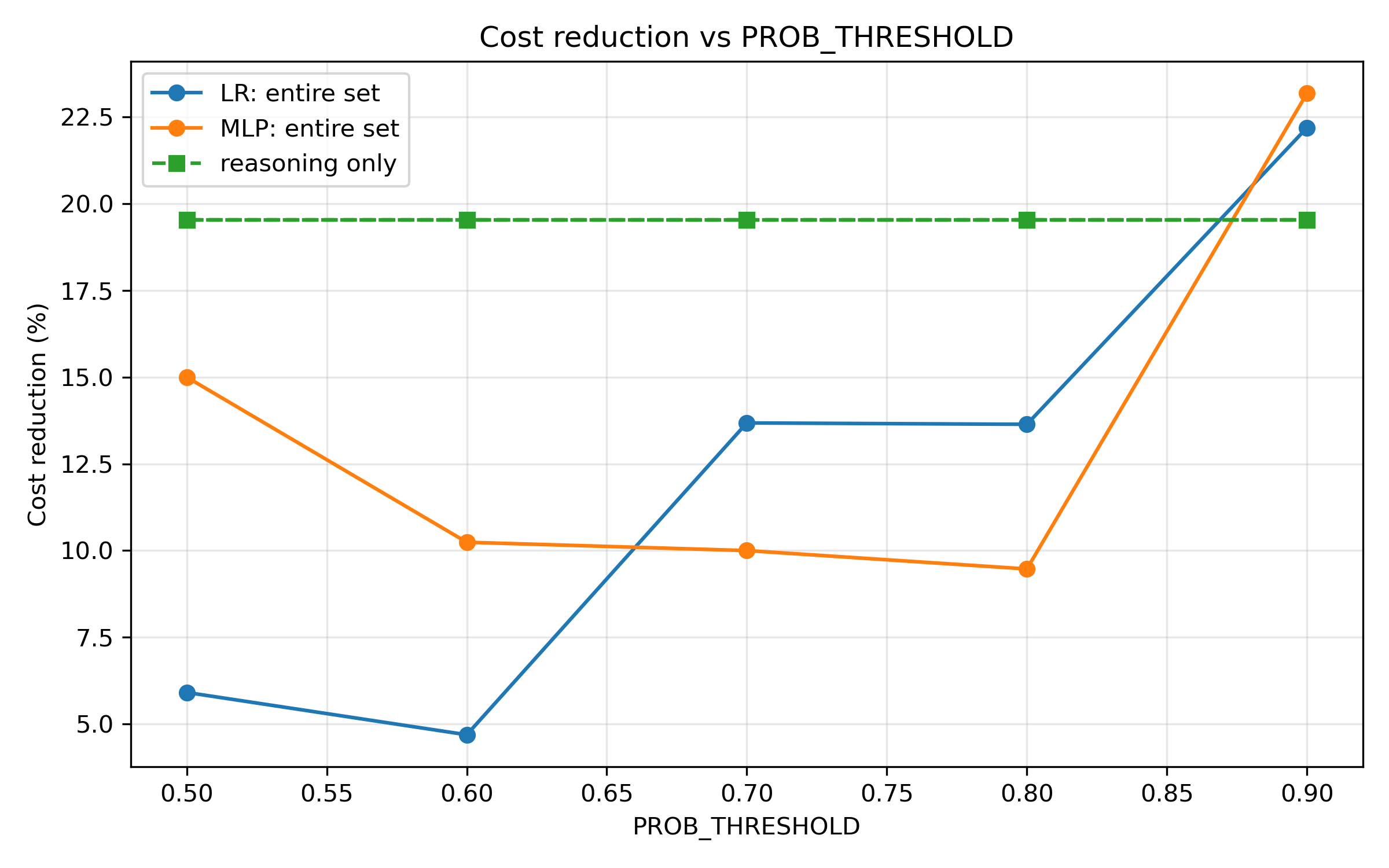}
\caption{IHCA prediction. Comparison of return (cost reduction) achieved by LR correction, MLP correction, and the
baseline Reasoning Only approach}
 \label{fig:LR_MLP_costs_vs_threshold_mimic_3_timestamped_HCA}
\end{figure}

\begin{figure}[H]
\centering
\includegraphics[scale=0.9]{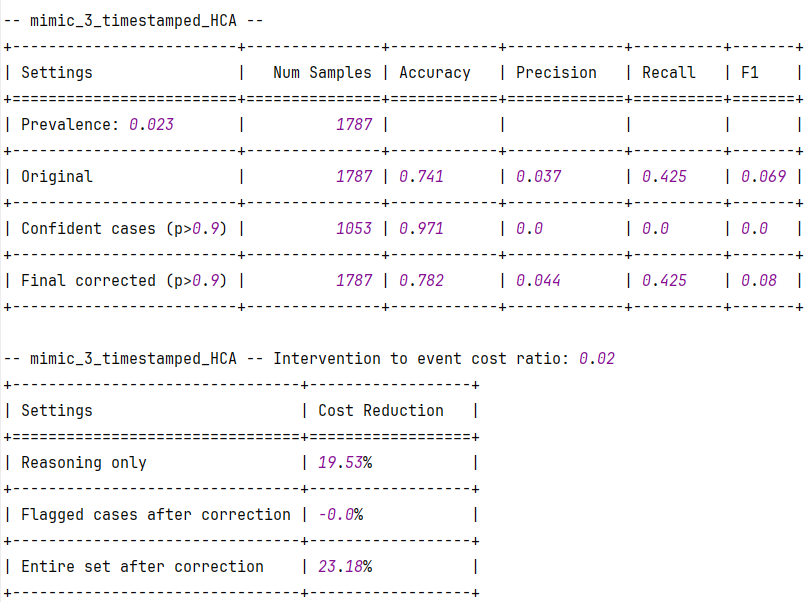}
\caption{IHCA prediction. Configuration that achieves the highest cost reduction}
 \label{fig:mimic3_timestamped_HCA_best_Prob_Thresh_MLP}
\end{figure}

Results are demonstrated in Figures \ref{fig:LR_metrics_vs_threshold_mimic_3_timestamped_HCA}, \ref{fig:MLP_metrics_vs_threshold_mimic_3_timestamped_HCA}, \ref{fig:LR_MLP_costs_vs_threshold_mimic_3_timestamped_HCA} and \ref{fig:mimic3_timestamped_HCA_best_Prob_Thresh_MLP} following the same agenda as in previous examples. In this case we can see that the MLP corrector outperforms LR one and both outperform the Reasoning Only case, that confirms importance of using CM in the rare event outcome prediction. 

\section{Results discussion}

The proposed LPCORP framework was evaluated on three representative rare-event prediction tasks spanning consumer finance and healthcare. For each dataset, we compared two correction models, Logistic Regression (LR) and Multilayer Perceptron (MLP), over a range of correction probability thresholds and evaluated both predictive performance and expected cost reduction.

The experiments demonstrate that the correction stage can substantially improve the practical value of reasoning-based predictions. Although the reasoning model already provides meaningful predictions, applying a lightweight correction model frequently increases precision while maintaining competitive recall, resulting in lower expected operational cost. The optimal correction model, however, depends on the application domain.

This observation supports the hypothesis that LLM reasoning contains information useful not only for generating predictions but also for estimating the reliability of those predictions.

For the Consumer Finance Complaint dataset, the MLP correction model consistently outperformed LR in terms of expected cost reduction. The best performance was obtained at a correction probability threshold of 0.8. This result suggests that the nonlinear decision boundary learned by MLP better captures patterns associated with incorrect LLM conclusions in consumer complaint narratives. Nevertheless, the improvement over the Reasoning Only baseline was relatively modest, indicating that the reasoning model itself already performs well on this task.

For the 30-day hospital readmission prediction task, Logistic Regression achieved the highest return (cost reduction), outperforming both the MLP correction model and the Reasoning Only baseline. The best performance was obtained with a probability threshold of 0.5. This observation indicates that the correction problem in this dataset is nearly linearly separable in the TF--IDF feature space and therefore benefits from the simpler LR model.

The most challenging task was timestamped in-hospital cardiac arrest (IHCA) prediction, where the prevalence of positive events was only approximately 2.3\%. In this highly imbalanced setting, both correction models improved upon the Reasoning Only baseline, with MLP producing the highest overall cost reduction. These results demonstrate that the proposed two-stage framework remains effective even under extreme class imbalance and that the correction stage is particularly valuable when the reasoning model alone generates a relatively large number of costly errors.

An important observation across all experiments is that the optimal correction probability threshold is dataset dependent. Although thresholds between 0.5 and 0.9 were evaluated, no single threshold consistently produced the best results. Consequently, the threshold should be selected using a validation dataset according to the application's objective, such as maximizing expected cost reduction or achieving a desired precision--recall trade-off.

Overall, the experiments show that a lightweight correction model can effectively complement reasoning-based prediction. Depending on the application, either LR or MLP may be preferable, but in all healthcare examples the correction stage reduced the expected operational cost relative to relying solely on the reasoning model. These findings suggest that combining reasoning-enhanced prediction with confidence-based correction provides a practical and computationally efficient approach for low-prevalence prediction problems.

\section{Conclusion}

In this study, we introduced LPCORP (Low-Prevalence CORrector for Prediction), a two-stage framework designed to address classification problems in which the positive class represents a rare event and the data are highly imbalanced. Rather than relying solely on the final prediction produced by a reasoning model, LPCORP exploits the intermediate reasoning itself. The experimental results suggest that the reasoning process generated by LLMs contains rich semantic signals, latent decision patterns, uncertainty cues, and contextual relationships that are not fully reflected in the final conclusion. Even when the reasoning model produces an incorrect prediction, these intermediate representations often preserve information that enables a lightweight correction model to identify and correct the error.

The proposed framework first generates reasoning and a prediction from narrative text using a reasoning-enabled LLM and then applies a lightweight correction model to determine whether the predicted outcome should be retained or reversed. In this study, we evaluated both Logistic Regression (LR) and a Multilayer Perceptron (MLP), demonstrating that the preferred correction model depends on the application domain while remaining computationally inexpensive and straightforward to deploy.

We presented a simplified mathematical formalization of the correction process and evaluated the method on multiple real-world prediction tasks from healthcare and consumer finance. Across these applications, LPCORP consistently improved the practical utility of reasoning-based prediction. The correction stage increased precision while maintaining competitive recall, leading to improved cost reduction compared with using the reasoning model alone. Furthermore, optimizing the correction probability threshold produced additional gains, illustrating the importance of application-specific threshold selection. Under realistic cost assumptions, the proposed framework achieved substantial operational benefit, including more than 40\% expected cost reduction in the early warning predictions.

Although the present study focused on binary rare-event prediction, the proposed framework is general and can be combined with different reasoning models and correction architectures. Future work will investigate larger reasoning models, richer correction models, and additional medical applications, including diabetes outcomes, ICU transfer prediction, and other clinical decision-support tasks. We also plan to investigate the interpretability of reasoning-derived features and better understand which semantic patterns within LLM-generated reasoning contribute most to successful outcome correction.
\\\\
\subsubsection*{References}

\noindent [1] He, H., \& Garcia, E. (2009). Learning from Imbalanced Data. IEEE Transactions on Knowledge and Data Engineering.

\noindent [2] Chawla, N. V., Bowyer, K., Hall, L., \& Kegelmeyer, W. (2002). SMOTE: Synthetic Minority Over-sampling Technique. Journal of Artificial Intelligence Research.

\noindent [3] Saito, T., \& Rehmsmeier, M. (2015). The Precision-Recall Plot is More Informative than the ROC Plot When Evaluating Binary Classifiers on Imbalanced Datasets. PLoS ONE.

\noindent [4] Niculescu-Mizil, A., \& Caruana, R. (2005). Predicting Good Probabilities with Supervised Learning. ICML.

\noindent [5] Christodoulou, E. et al. (2019). A Systematic Review Shows No Performance Benefit of Machine Learning Over Logistic Regression for Clinical Prediction. Journal of Clinical Epidemiology.

\noindent [6] Lee, H-Y. et al. (2024). Prediction of in-hospital cardiac arrest in the intensive care unit using a machine-learning multimodal approach. JMIR Medical Informatics.

\noindent [7] Shaffiee Haghshenas, S. et al. (2025). Role of Artificial Intelligence in Critical Care Medicine: A Literature Review. Cureus.

\noindent [8] Schmidt, M., et al. (2023). Anomaly Detection for Predictive Maintenance Using Unsupervised Learning. IEEE Transactions on Industrial Informatics.

\noindent [9] Dal Pozzolo, A., Boracchi, G., Caelen, O., Alippi, C., \& Bontempi, G. (2018). Credit card fraud detection: a realistic modeling and a novel learning strategy. IEEE Transactions on Neural Networks and Learning Systems.

\noindent [10] Lala, J., et al. (2023). Early Prediction of Extreme Weather Events Using Deep Learning. Nature Communications.

\noindent [11] An, S., Kim, J., Choi, G., Jang, H., \& Ahn, K. (2024). The effect of rare events on information-leading role: evidence from real estate investment trusts and overall stock markets. Humanities and Social Sciences Communications, 11, 1628.

\noindent [12] Lightman, H. et al. (2023). Let’s Verify Step by Step. https://arxiv.org/abs/2305.20050

\noindent [13] Baaj, I. et al. (2024). Synergies between Machine Learning and Reasoning. International Journal of Approximate Reasoning, 171, 109206.

\noindent [14] J. Brownlee, Cost-Sensitive Learning for Imbalanced Classification, 2020, https://machinelearningmastery.com/cost-sensitive-learning-for-imbalanced-classification

\noindent [15] https://huggingface.co/deepseek-ai/DeepSeek-R1-Distill-Llama-8B

\noindent [16] T. Fawcett, F Provost, Adaptive fraud detection. Data mining and knowledge discovery, 1997 Springer

\noindent [17] C. Elkan, The Foundations of Cost-Sensitive Learning, Proceedings of the Seventeenth International Joint Conference on Artificial Intelligence (IJCAI’01), 2001.

\noindent [18] D. Hand Evaluating diagnostic tests: The area under the ROC curve and the balance of errors, Stat Med., 2010 Jun 30;29(14):1502-10

\noindent [19] Charlotte Haendler et al. The Hidden Costs of Financial Services: Consumer Complaints and Financial Restitution. SMU Cox School of Business Research Paper No. 25-11, 2025

\noindent [20] Enforcement by the numbers. Consumer Financial Protection Burau, 2025, consumerfinance.gov

\noindent [21] Voiso call-center cost benchmarks; Industry reports on U.S. customer service labor costs.

\noindent [22] Audrey J. Weiss et al. Overview of Clinical Conditions With Frequent and Costly Hospital Readmissions by Payer, 2018, STATISTICAL BRIEF \#278, July 2021

\noindent [23] M. McDermott et al., Reproducibility in machine learning for health research: Still a ways to go. Science Translational Medicine, 13(586), eabb1655 (2021).

\noindent [24] B. Wellner, et al., Predicting Unplanned Transfers to the Intensive Care Unit: A Machine Learning Approach Leveraging Diverse Clinical Elements. JMIR Med Inform 2017 in Vol 5, No 4 (2017): Oct-Dec

\noindent [25] Andersen LW et al., In-Hospital Cardiac Arrest: A Review, JAMA The Journal of the American Medical Association, 2019

\noindent [26] Holmberg, Mathias J.; Ross, Catherine E.; Fitzmaurice, Garrett M. et al. / Annual incidence of adult and pediatric in-hospital cardiac arrest in the United States. In: Circulation: Cardiovascular Quality and Outcomes. 2019; Vol. 12, No. 7.

\noindent [27] Paul S Chan et al. Readmission Rates and Long-Term Hospital Costs Among Survivors of In-Hospital Cardiac Arrest. Circ Cardiovasc Qual Outcomes. 2014 Oct 28;7(6):889–895.

\noindent [28] Abdulla A. Damluji et al. Health Care Costs After Cardiac Arrest in the United States. Circulation: Arrhythmia and Electrophysiology, Volume 11, Number 4, 2018

\noindent [29] Vinodh B Nanjayya et al. Actual Cost of Extracorporeal Cardiopulmonary Resuscitation: A Time-Driven Activity-Based Costing Study. Crit Care Explor. 2024 Jul 3;6(7)

\noindent [30] Bonafide CP et al. Cost-benefit analysis of a medical emergency team in a children’s hospital. Pediatrics. 2014;134(2):235–241.

\noindent [31] McGaughey J et al. Outreach and Early Warning Systems for the prevention of ICU admission and death of critically ill adult patients on general hospital wards. Cochrane Database Syst Rev. 2007; CD005529.

\noindent [32] Jones DA et al Rapid-response teams. N Engl J Med. 2011;365(2):139-146.
\\\\
\textit{Computations were performed using a single NVIDIA RTX 6000 GPU with 48 GB at Division of Endocrinology, Mass General Brigham, Boston, MA}

\end{document}